\documentclass[10pt,twocolumn,letterpaper]{article}

\usepackage{cvpr}
\usepackage{times}
\usepackage{epsfig}
\usepackage{graphicx}
\usepackage{amsmath}
\usepackage{amssymb}
\usepackage{multirow}
\usepackage{subfigure}
\usepackage{color}
\usepackage{booktabs}

\usepackage{float}
\usepackage{caption}

\usepackage[pagebackref=true,breaklinks=true,letterpaper=true,colorlinks,bookmarks=false]{hyperref}

\cvprfinalcopy 

\def\cvprPaperID{3959} 

\ifcvprfinal\fi
\begin{document}
\pagenumbering{gobble} 

\title{Modulating Image Restoration with Continual Levels \\via Adaptive Feature Modification Layers}

\author{Jingwen He$^{1,}$\thanks{The first two authors are co-first authors. (e-mail: jw.he@siat.ac.cn; chao.dong@siat.ac.cn).} \quad Chao Dong$^{1,}$\footnotemark[1] \quad Yu Qiao$^{1,2,}$\thanks{Corresponding author (e-mail: yu.qiao@siat.ac.cn).}\\
	$^1$ShenZhen Key Lab of Computer Vision and Pattern Recognition, \, SIAT-SenseTime Joint Lab, \\ 
	Shenzhen Institutes of Advanced Technology, Chinese Academy of Sciences, China\\
	$^2$The Chinese University of Hong Kong\\
}

\maketitle

\begin{abstract}

\vspace{-1em}
In image restoration tasks, like denoising and super-resolution, continual modulation of restoration levels is of great importance for real-world applications, but has failed most of existing deep learning based image restoration methods. Learning from discrete and fixed restoration levels, deep models cannot be easily generalized to data of continuous and unseen levels. This topic is rarely touched in literature, due to the difficulty of modulating well-trained models with certain hyper-parameters. We make a step forward by proposing a unified CNN framework that consists of few additional parameters than a single-level model yet could handle arbitrary restoration levels between a start and an end level. The additional module, namely AdaFM layer, performs channel-wise feature modification, and can adapt a model to another restoration level with high accuracy. By simply tweaking an interpolation coefficient, the intermediate model -- AdaFM-Net could generate smooth and continuous restoration effects without artifacts. Extensive experiments on three image restoration tasks demonstrate the effectiveness of both model training and modulation testing. Besides, we carefully investigate the properties of AdaFM layers, providing a detailed guidance on the usage of the proposed method.
\end{abstract}

\section{Introduction}

Deep learning methods have achieved great success in image restoration tasks, such as denoising, super-resolution, compression artifacts reduction, etc \cite{lim2017enhanced, ledig2017photo, dong2014learning, Dong_2015_ICCV, zhang2017beyond}. However, there still exists a large gap of restoration performance between research environment mand real-world applications. In this work, we focus on two main issues that prevent CNN based restoration methods from wide usages. 

\begin{figure}[]
\vspace{-0.5em}
\centering
\includegraphics[scale=0.21]{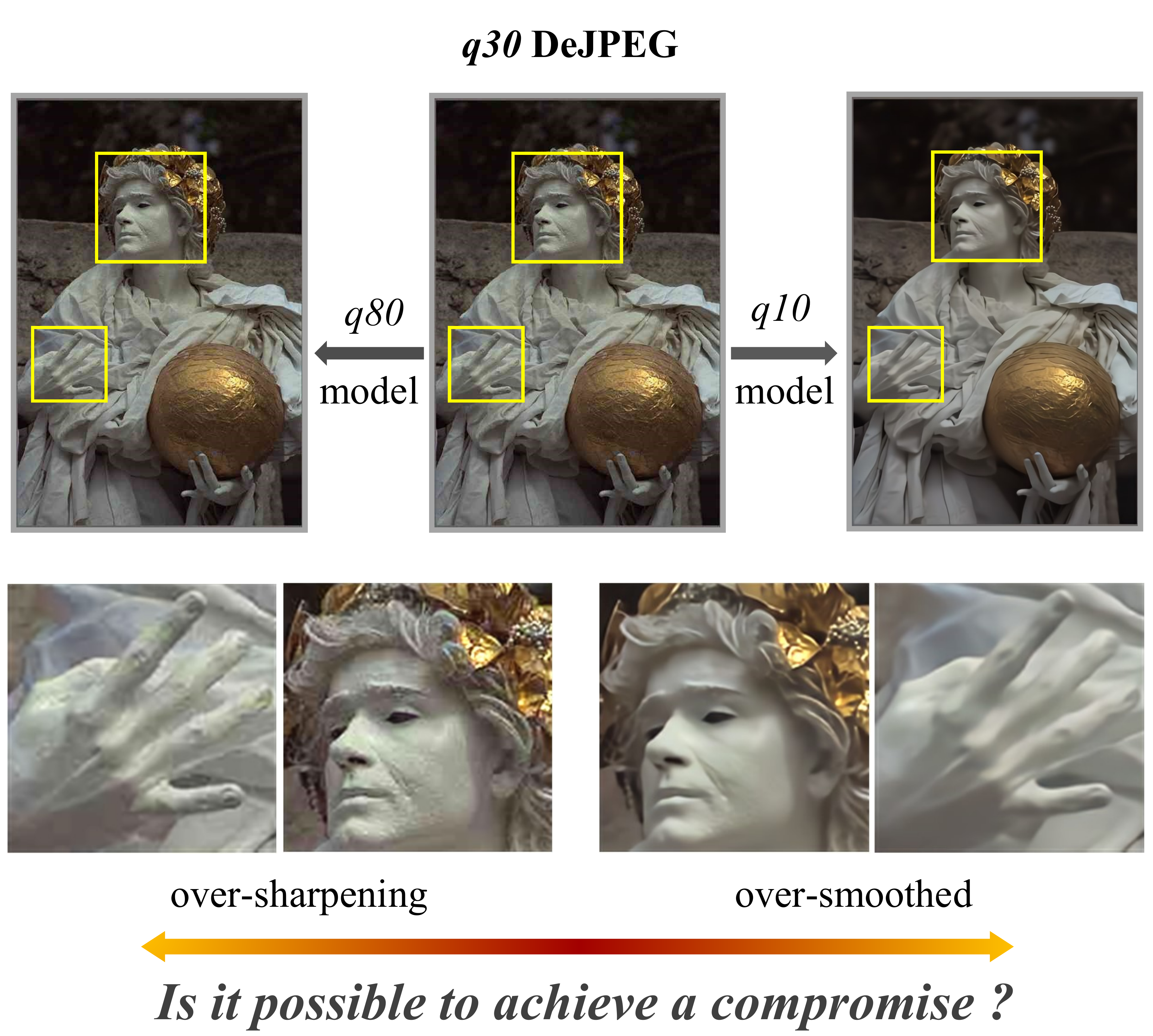}
\vspace{-0.5em}
\captionsetup{font={small}}
\caption{Applying \textit{q10} or \textit{q80} DeJPEG model on images (LIVE1 \cite{1709988}) with degradation \textit{q30} tends to produce either over-sharpening (left) or over-smoothed (right) images.}
\vspace{-2em}
\label{fig:problem}
\end{figure}

First, the degradation levels of real-world images are generally continuous, such as JPEG quality $q27$ and $q34$. On the other hand, the deep restoration models are usually trained with discrete and fix levels (e.g., $q20$, $q30$). Applying models with mismatched restoration levels tends to produce either over-sharpening or over-smoothed images, as shown in Figure~\ref{fig:problem}\footnote{DeJPEG is also known as JPEG deblocking and compression artifacts reduction.}. A straightforward solution is to train a sufficiently large model to handle all degradation levels. However, regardless of the computational burden, this general model is not optimal for each individual level. When we want to slightly adjust the output effects, we have to retrain a new model by refining the model structure, parameters or (and) loss functions, which is a tedious procedure with unpredictable results.

\begin{figure}[]
	\centering
	\includegraphics[scale=0.21]{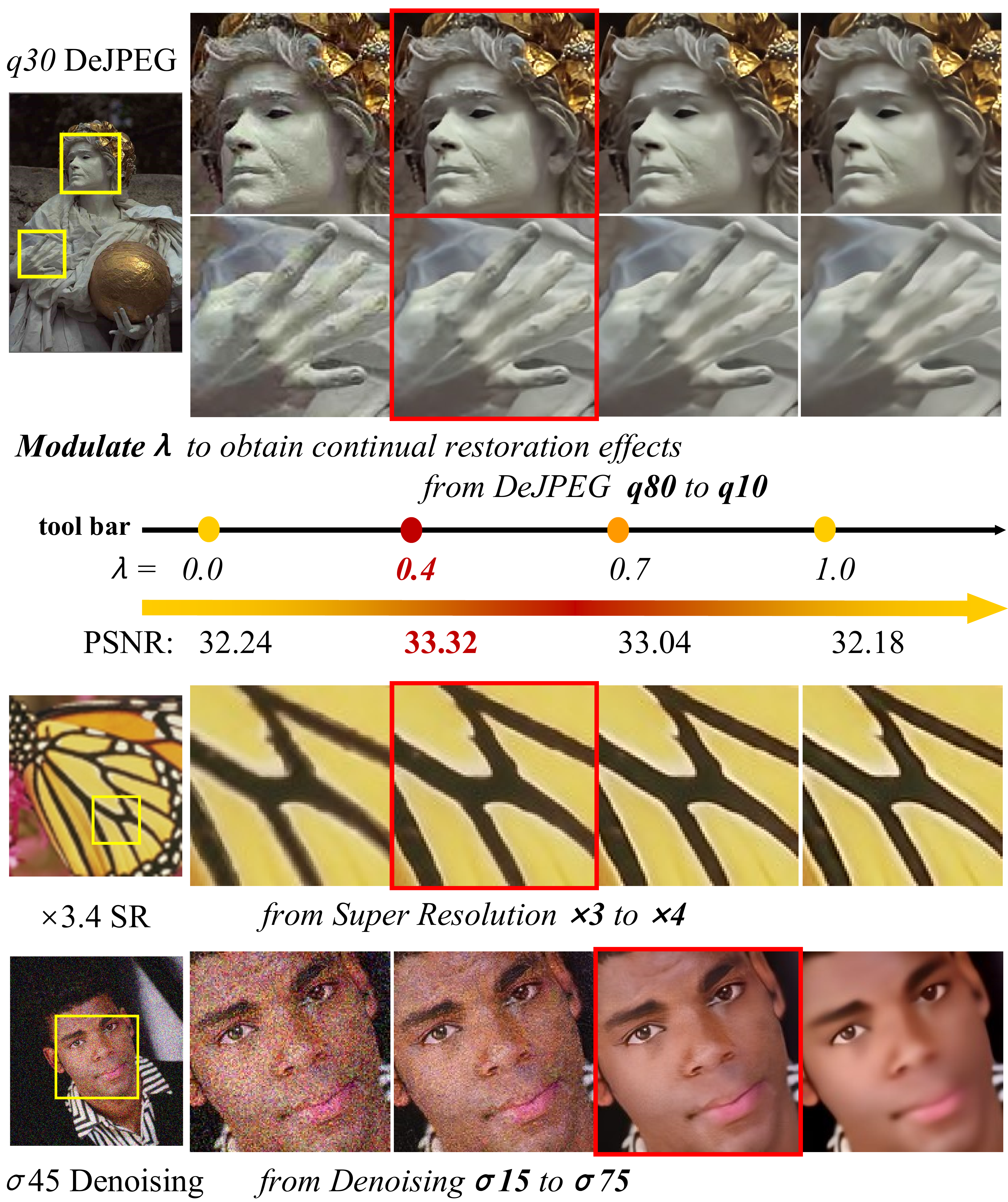}
	\vspace{-0.5em}
	\captionsetup{font={small}}
	\caption{We can modulate the tool bar to obtain continual restoration effect in DeJPEG, Super Resolution and Denoising.}
	\vspace{-2em}
	\label{fig:solution}
\end{figure}

Second, in industrial and commercial scenarios (e.g., human-interactive softwares), it is often necessary to consecutively modulate the restoration strength/effect to meet different requirements. For example, the users always expect a tool bar to flexibly adjust the restoration level, as shown in Figure \ref{fig:solution}. However, current deep models are trained on fixed degradation levels, and contain no hyper-parameters for users to change the final results. 

To fill in the gaps, our goal is to achieve arbitrary-level image restoration and continual model modulation in a unified CNN framework. More formally, the task is to deal with images of degradation levels between a ``start'' level and an ``end'' level in a user controllable manner. To facilitate practical usages, we should avoid building a very large model or model zoo, and prevent another training stage at test time. In other words, the solution should contain a small amount of additional parameters and allow continual tuning of parameters in testing. 

This task is non-trivial and rarely studied in literature. Perhaps the most relevant topic to modifying the network outputs is arbitrary style transfer. Specifically, we can treat different levels of degradation as different kinds of styles. A representative approach is the Conditional Instance Normalization (IN) \cite{dumoulin2017learned}, which allows users to mix up different styles by tuning IN parameters. Nevertheless, image restoration has higher and finer request on the output image quality. Directly applying Conditional IN in image restoration could produce obvious and large-scale artifacts in the output image (see Figure \ref{fig:baselines}). Another similar concept is domain adaptation, which generally appears in high-level vision problems (e.g., image classification and object detection). It adapts/transfers the model trained on the source domain to the target domain. However, domain adaptation cannot easily generalize to unseen data, thus is not appropriate to address our problem.

In this work, we present a simple yet effective approach that for the first time enables consecutive modulation of the restoration strength with little computation cost. This approach stems from the observation that filters among networks of different restoration levels are similar at patterns while varying on scales and variances. Furthermore, the model outputs could change continuously by modulating the statistics of features/filters. The proposed framework is built upon a novel Adaptive Feature Modification (AdaFM) layer that modifies the middle-layer features with depth-wise convolution filters. In practice, we first train a standard restoration CNN for the start level, and then insert AdaFM layers and optimize it to the end level. After the training stage, we fix the CNN parameters, and interpolate the filters of AdaFM layers according to testing restoration level. By tuning a controlling coefficient (ranging from 0 to 1), we can interactively and consecutively manipulate the restoration results/effects. Note that we only need to train the CNN and AdaFM layers once, and no further training is required in the test time.

To ensure the output quality, we demonstrate that the model with AdaFM layers achieves comparable performance to the single-level image restoration network in both start and end level. Then, we show that the modulated-network outputs are noise-free with consecutive restoration effects (see Figure \ref{fig:solution}). Besides, we also examine the properties of AdaFM layers –- complexity, range and direction, providing a detailed instruction on the usage of the proposed method. Notably, the added AdaFM layers contribute to less than 4\% parameters of the CNN model yet achieves excellent modulation performance.

\section{Related Work}

The proposed Adaptive Feature Modification (AdaFM) layers are inspired by the recent normalization methods in deep CNNs, thus we give a brief review of these works. Normalization has been demonstrated effective in facilitating training very deep neural networks. 
The most representative method is batch normalization (BN) \cite{Ioffe2015BatchNA} that is proposed to address the problem of Internal Covariate Shift in the training process. In particular, BN layer normalizes the output of each neuron using the mean and variance of each batch calculated during the feed-forward process. 
Later on, Dmitry Ulyanov et al. \cite{DBLP:journals/corr/UlyanovVL16} achieved significant improvement in style transfer by replacing all the BN layers with their proposed instance normalization (IN) layers. The core idea is to normalize the features based on the statistics across the spatial dimensions of each sample instead of each batch. 
Recently, several alternative normalization methods have been proposed, such as instance weight normalization \cite{salimans2016weight}, layer normalization \cite{Ba2016LayerN}, group normalization (GN) \cite{wu2018group} and etc.
The spatial feature transformation (SFT) layer proposed by Wang~et al. \cite{wang2018recovering} further extends the normalization operation to a more general spatial-variant transformation. Specifically, they apply a feature spatial-wise transformation on the feature maps according to the semantic segmentation priors. This approach indeed helps generate more realistic textures compared with those popular GAN-based methods. We will compare the proposed AdaFM layer with BN and SFT layers in Section~\ref{sec:AdaFM}.

Furthermore, recent works show that BN and IN have the ability to adapt the model to a different domain with little computation cost.
Specifically, Li et al. \cite{Li2017RevisitingBN} propose AdaBN (Adaptive Batch Normalization) to alleviate domain shifts, and show that AdaBN is effective for domain adaptation task by re-computing the statistics of all BN layers across the network. Huang et al. \cite{huang2017arbitrary} show that instance normalization (IN) can perform as style normalization by aligning the mean and variance of content features with those style features. In such way, they realize arbitrary style transfer at test time. Moreover, Dumoulin et al. \cite{dumoulin2017learned} extended IN to enable multiple style transfer by learning different sets of parameters in normalization layers while the convolution parameters are shared. Our method is different from these works in that 1) the proposed AdaFM layer is independent of either batch or instance samples, 2) the filter size and position of AdaFM layers are flexible, indicating that AdaFM is beyond a normalization operation, 3) the interpolation property of AdaFM layers could achieve continual modulation of restoration levels, which has not been revealed before.

\section{Method}

\subsection{Problem Formulation.}
The problem of consecutive modulation of restoration levels can be formulated as follows.
Suppose we have a ``start'' restoration level -- $L_a$ and an ``end'' restoration level -- $L_b$, the objective is to construct a deep network to handle images with arbitrary degradation level $L_c$ ($L_a\le L_c \le L_b$). 
Our solution pipeline consists of two stages -- \textit{model training} and \textit{modulation testing}. In model training, we train a basic model and an adaptive model that could deal with level $L_a$ and $L_b$, respectively. While in modulation testing, we propose a new network that can realize arbitrary restoration effects between level $L_a$ and $L_b$ by modulating certain hyper-parameters. 
In the following sections, we first show two important observations that inspire our method. Then we propose the AdaFM layer and compare it with BN \cite{Ioffe2015BatchNA} and SFT \cite{wang2018recovering}. At last, we describe how to use AdaFM layers in model training and modulation testing.

\subsection{Observation}

\begin{figure}[t]
\centering
\includegraphics[scale=0.32]{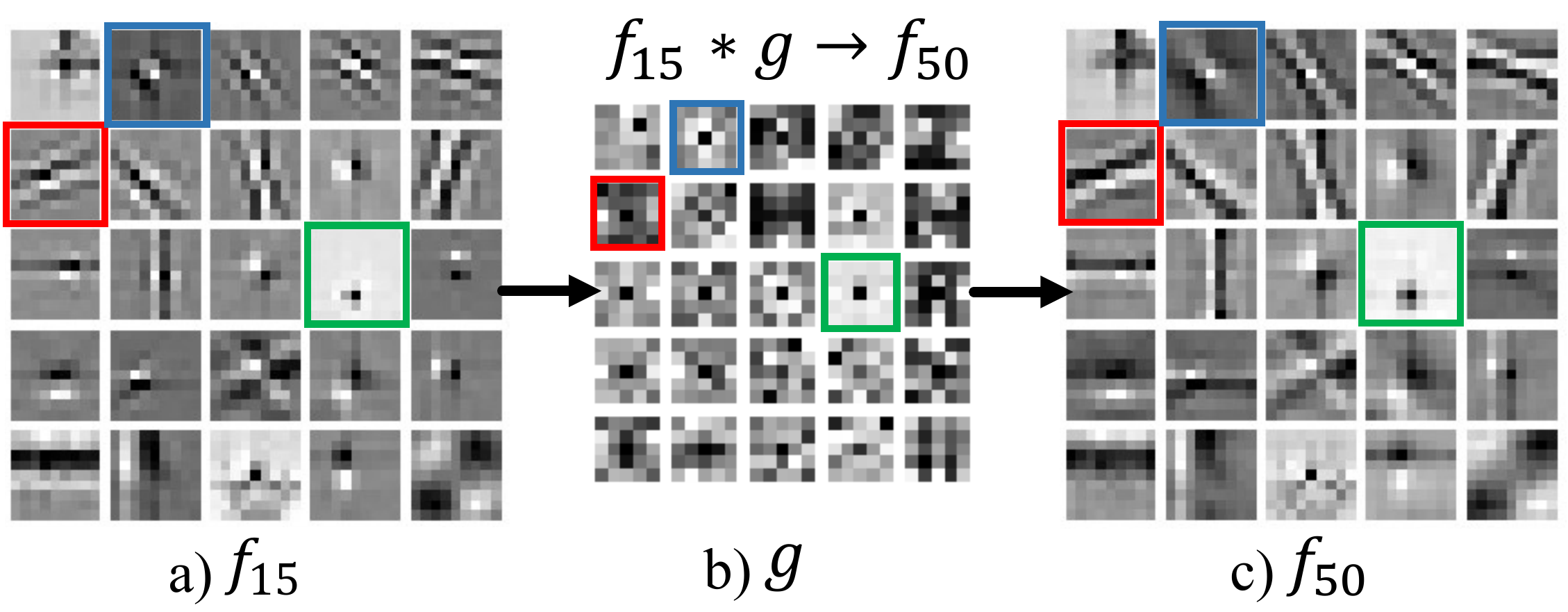}
\captionsetup{font={small}}
\vspace{-0.5em}
\captionsetup{font={small}}
\caption{Filter visualization.}
\vspace{-1.5em}
\label{filters_noise}
\end{figure}

\textbf{Observation 1.} We find that the learned filters of restoration models trained with different restoration levels are pretty similar at visual patterns, but their weights have different statistics (e.g., mean and variance). An example is shown in Figure \ref{filters_noise}, the filter $f_a$ of level $L_a$ is like a 2-D Gaussian filter, then the corresponding filter $f_b$ finetuned from level $L_a$ to level $L_b$ will also look like a Gaussian filter but with different mean and variance. 
We use the Gaussian Denoising problem for illustration. The start level is noise level $\sigma=15$, and the end level is $\sigma=50$. We adopt a simple and standard CNN structure ARCNN \cite{Dong_2015_ICCV} to do the experiments. We first learn the model with noise level $\sigma=15$ and obtain ARCNN-15, then finetune the network on $\sigma=50$ to obtain ARCNN-50. The first layer filters of these two models are visualized in Figure \ref{filters_noise}. In the first glance, these filters look similar with only slight differences. Their mean cosine distance between the corresponding filters is 0.12, indicating that they are very close to each other.
To further reveal their relationship, we use a filter to bridge the corresponding filters. Specifically, each filter $f_{15}$ in ARCNN-15 is convoluted with another filter $g$ to approximate the corresponding filter $f_{50}$ in ARCNN-50. According to the commutative law, we have $ (g*f_{15})*x = g*(f_{15}*x) $, where $*$ is convolution. Thus for each feature map $x$, the parameters of $g$ are optimized with 

\vspace{-1em}
\begin{equation}
  \min_{g}||f_{50}*x-g*(f_{15}*x)||^2.
\end{equation}
\vspace{-1em}

The above operation is equivalent to adding a depth-wise convolution layer after each layer of ARCNN-15, and finetuning the added parameters on the  $\sigma=50$ problem. When $g$ is of size $1\times1$, it is equal to a scaling and shift operation, changing the mean and variance of the original filter. We use the PSNR gap between their network outputs to show the fitting error. From Table \ref{table:kernel_size}, we can see that the value of fitting error decreases when the filter size of $g$ increases. The gap is already very small at $1\times1$, which demonstrates our primal assumption. The $5\times5$ filters are also visualized in Figure \ref{filters_noise}, where one can see the differences between $f_{15}$ and $f_{50}$. Similar experiments for super resolution and compression artifacts reduction are presented in the supplementary file. 

\begin{figure*}[t]
\centering
\vspace{-1em}
\includegraphics[scale=0.75]{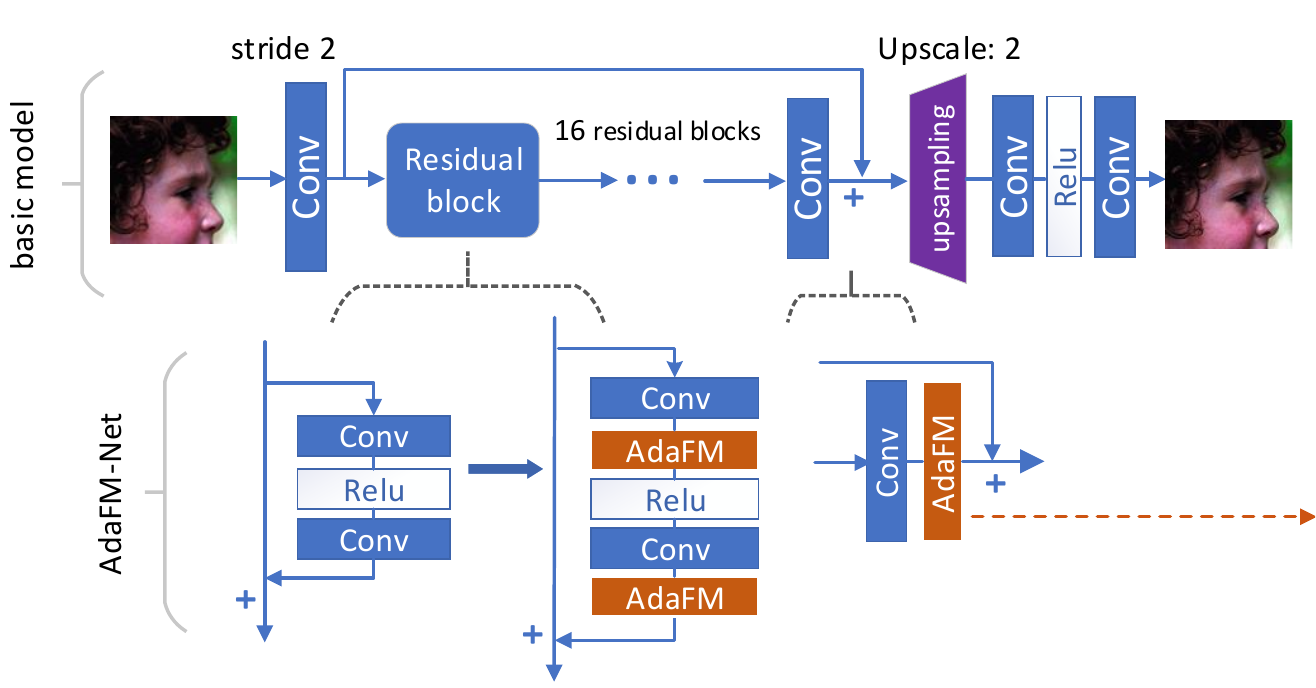}
\includegraphics[scale=0.3]{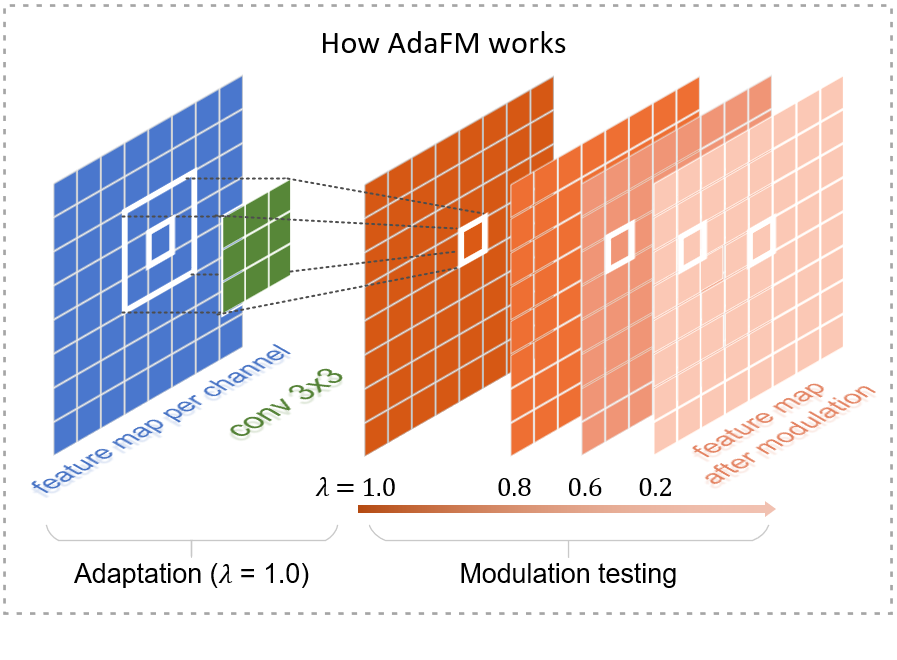}
\vspace{-1em}
\captionsetup{font={small}}
\caption{The left part presents the basic model and the AdaFM-Net. The right part shows how AdaFM works in the adaptation process and the modulation testing.}
\vspace{-1.5em}
\label{arch}
\end{figure*}

\textbf{Observation 2.} We find that the network output could change continuously by modulating the statistics of features/filters. As the filter $g$ is gradually updated by gradient descent, what if we control the updating process by interpolating the intermediate results? 
Specifically, we can obtain the intermediate filter $f_{mid}$ by the following function:

\vspace{-0.5em}
\begin{equation}
  f_{mid}=f_{15}+\lambda (g-I)*f_{15}, 0\le \lambda \le 1, 
\end{equation}
where $\lambda$ is an interpolation coefficient. When we modulate $\lambda$ gradually from 0 to 1, $f_{mid}$ will also change continuously from $f_{15}$ to $g*f_{15}$. After putting $f_{mid}$ back to the network, we find that the network output will also change continuously in visualization, as shown in Figure~\ref{fig:solution}. Detailed analysis can be found in Section~\ref{sec:test} and~\ref{sec:exp}.

\subsection{Adaptive Feature Modification}
\label{sec:AdaFM}
Inspired by the above observations, we propose a continual modulation method by introducing an Adaptive Feature Modification layer and the corresponding modulating strategy. The overall framework is depicted in Figure \ref{arch}. 

Our aim is to add another layer to manipulate the statistics of the filters, so that they could be adapted to another restoration level. As indicated in Observation 1, we can add a depth-wise convolution layer (or a group convolution layer with the group number equal to the number of feature maps) after each convolution layer and before the activation function (e.g., ReLU). We name the added layer as the Adaptive Feature Modification layer, which is formulated as
\vspace{-0.5em}
\begin{equation}
   AdaFM(x_i)=g_i * x_i + b_i, 0<i\le N,
\end{equation}
where $x_i$ is the input feature map and $N$ is the number of feature maps. $g_i$ and $b_i$ are the corresponding filter and bias, respectively. It is worth noting that $g_i$ depends on the degradation level of input images. To further understand its behaviour, we compare the proposed layer with batch normalization (BN) \cite{Ioffe2015BatchNA} and spatial feature transformation (SFT) \cite{wang2018recovering} layers.

\textbf{Comparison with BN layer.} When we set the filter size of $g_i$ to $1\times1$, the feature modification reduces to a normalization operation. Note that BN \cite{Ioffe2015BatchNA} is also put directly after the convolution layer. We compare it with BN as
\vspace{-0.5em}
\begin{equation}
\vspace{-0.5em}
   AdaFM(x_i)=g_i x_i + b_i, BN(x_i)=\gamma (\dfrac{x_i-\mu}{\sigma})+\beta,
\end{equation}
where $\mu,\beta$ are the mean and standard deviation of an input batch, $\gamma,\beta$ are affine parameters. The $1\times1$ AdaFM filter performs similar to BN without using the batch information. 
As a special case, we can also use BN to do feature modification and finetune $\gamma,\beta$ as $g_i$, $b_i$. Experiments show that using BN achieves almost the same results as the $1\times1$ AdaFM filter.

\textbf{Comparison with SFT layer.} When the filter size of $g$ is as large as the feature map, it will perform spatial feature transform as SFT layer \cite{wang2018recovering}. The formulations are shown as 
\vspace{-0.5em}
\begin{equation}
   AdaFM(x_i)=g_i \odot x_i + b_i, SFT(x_i)=\gamma \odot x_i+\beta,
\end{equation}
where $\gamma,\beta$ are affine parameters. AdaFM and SFT layer share the same function, but different on the parameters. Specifically, $\gamma,\beta$ are calculated from another sub-network based on an additional prior, while $g_i,b_i$ are directly learned with the network.

\subsection{Model Training}
\label{sec:train}
In this subsection, we discuss how to utilize the proposed AdaFM layer for model training. The entire model, namely AdaFM-Net, consists of a basic network and the AdaFM layers. First, we train the basic network $N_{bas}^a$, which can be any standard CNN model, for the start restoration level $L_a$. Then we insert AdaFM layers to $N_{bas}^a$ and form the AdaFM-Net $N_{ada}$. By fixing the parameters of $N_{bas}^a$, we optimize the parameters of AdaFM layers on the end level $L_b$. Experiments demonstrate that by only finetuning the AdaFM layers, the model $N_{ada}^b$ could achieve comparable performance with a basic model $N_{bas}^b$ trained from scratch on level $L_b$. As the AdaFM-Net is optimized from $L_a$ to $L_b$, we name this process as adaptation, and use \textit{adaptation accuracy} to denote its performance. Specifically, we can use the PSNR distance between PSNR of $N_{ada}^b$ and $N_{bas}^b$ as the measurement of adaptation accuracy.
There are three factors that affect the adaptation accuracy -- filter size, direction, and range.

(1) For filter size, a larger filter size or more parameters will lead to better adaptation accuracy. We try filter size from $1\times 1$ to $7\times 7$. From convergence curves shown in Figure~\ref{kernel_size_figure}, we find that $3\times 3$ performs much better than $1\times 1$ while $7\times 7$ is only comparable to $5\times 5$. Further increasing the filter size could not continuously improve the performance. 
(2) For direction, different restoration levels have different degrees of difficulty for the same network. Then should we modulate the model from an easy level to a hard level or the opposite direction? Experimentally, we find that from easy to hard is a better choice (see Section~\ref{sec:direction}). 
(3) For range, the smaller of the range/gap $|L_b-L_a|$, the better the adaptation accuracy. For example, in super resolution problem, transferring the filters from $\times 2$ to $\times 3$ is easier than from $\times 2$ to $\times 4$. In Section~\ref{sec:exp}, we conduct numerous experiments to choose the best range for super-resolution, denoising and compression artifacts reduction.

\subsection{Modulation testing}
\label{sec:test}
After the training process, we discuss how to modulate the AdaFM layers according to degradation level at test time. As the features remain the same after convolution with an identity filter, we initialize AdaFM layers with identity filters $I$ and zero biases, which is regarded as the start point of AdaFM layers. Based on Observation 2, we can linearly interpolate the parameters of AdaFM layers as
\begin{equation}
  g_i^*=I+\lambda (g_i-I), b_i^* =\lambda b_i,0<i\le N,
\end{equation}
where $g_i^*,b_i^*$ are the filter and bias of the interpolated AdaFM layers, $\lambda (0\le \lambda \le 1)$ is the interpolation coefficient determined by the degradation level of input image. After adding the interpolated AdaFM layers back to the basic network $N_{bas}^a$, we can get the AdaFM-Net $N_{ada}^c$ for a middle level $L_c(L_a\le L_c \le L_b)$. The effects of changing the coefficient $\lambda$ from 0 to 1 are shown in Figure \ref{fig:solution}, \ref{fig:baselines}, where the output effects change continuously along with $\lambda$. 

Interestingly, we find that the interpolated network could fairly deal with any restoration level $L_c$ between level $L_a$ and $L_b$ by adjusting the coefficient $\lambda$, which behaves like a strength controller in traditional methods. Experimentally, we find that the relationship between the coefficient $\lambda$ and restoration level $L_c$ can be formulated/approximated as a polynomial function:
\vspace{-1em}
\begin{equation}
  \lambda =f( L_c )=\sum_{j=0}^M w_j L_c ^j,
  \vspace{-0.5em}
\end{equation}
where $M$ is the order and $\{w_j\}_0^M$ are coefficients. To fit this polynomial function, we need to determine at least $M$ points – $\{L_c^i, \lambda^i\}_{i=0}^{M}$. Specially, the start point is $\{ L_c^0=L_a , \lambda^0=0 \} $ and the end point is $\{\lambda^M=1, L_c^M=L_b \}$. Furthermore, we require a test set with degraded images and ground truth to measure the adaptation accuracy. For a middle level $L_c^i$ , we use the test images of level $L_c^i$ as inputs. By adjusting the coefficient $\lambda $, the AdaFM-Net could generate a series of outputs. We select the $\lambda$ that achieves the highest PSNR (evaluated on the test set) as the best coefficient, recorded as $\lambda^i $ for $L_c^i$. It is worth noticing that the modulation process and curve fitting require no additional training.

Extensive experiments show that the fitting curve varies a lot with ranges and problems. Take compression artifacts reduction as an example. If the range is small, such as JPEG quality from $q80$ to $q50$, then the fitting function is linear (order $M=1$) as shown in Figure \ref{fig:modulation}. On the other hand, if the range is large, such as from $q80$ to $q10$, then we have to use a  curve (order $M=3$) for approximation. Similar trend is observed for denoising and super resolution (see details in Section~\ref{sec:modulation} and the supplementary file).

As an alternative choice, we can also use the piece-wise linear function for approximation. Actually, when the range is small enough, the relationship between $\lambda$ and $L_c$ is almost linear. We can train a set of AdaFM-Nets on middle levels $\{L_c^i\}$. For a given level $L_c$ ($L_c^i<L_c<L_c^{i+1}$), we can use the coefficient $\lambda=(L_c-L_c^i)/(L_c^{i+1}-L_c^i)$ to interpolate the AdaFM-Nets between $L_c^i$ and $L_c^{i+1}$. This strategy needs to train and store more AdaFM-Nets on middle levels, but the adaptation accuracy is comparably higher due to the small range. 

\section{Experiments}
\label{sec:exp}

\subsection{Experimental Set-up}

\textbf{Training settings.} 
We use the DIV2K \cite{agustsson2017ntire} dataset for all the image restoration tasks. The training data is augmented by horizontal flipping and 90-degree rotations. Following SRResNet \cite{ledig2017photo}, the mini-batch size is set to 16 and the HR patch size is 96 $\times$ 96. The L1 loss \cite{wang2018esrgan} is adopted as the loss function. For model training, the initial learning rate is set to $1\times10^{-4}$ and then decayed by a factor of 10 after  $5\times10^{5}$ iterations. We adopt the Adam \cite{Kingma2015AdamAM} optimizer with $\beta_{1}=0.9$, $\beta_{2}=0.999$. All models are built on the PyTorch framework and trained with NVIDIA 1080Ti GPUs.

\textbf{The structure of basic model.}
Based on the widely used SRResNet and DnCNN \cite{zhang2017beyond}, the basic model $N_{bas}$ adopts a general CNN structure that consists of a pair of down-sampling (convolution with stride 2) and up-sampling (pixelshuffle \cite{shi2016real} with upscaling factor 2) layers, 16 residual blocks, and several convolution layers. Specifically, the filter number is 64 and the filter size is $3\times3$ for all convolution layers. The residual block contains two convolution layers and a ReLU activation layer. The middle features are processed in a low-resolution (1/4 of the input size) space, while the output size remains the same as the input size. For super-resolution, we can upsample the LR image to the HR image size as SRCNN \cite{dong2014learning}. As shown in Table~\ref{tabel:basic}, the basic model achieves better PSNR results than SRResNet, DnCNN and ARCNN on super-resolution, denoising and compression artifacts reduction, respectively. As stated in Section~\ref{sec:train} and \ref{sec:test}, the basic model is also trained on different levels (as the baseline) to evaluate the performance of AdaFM-Nets. 

\textbf{The position of AdaFM layers.}
As indicated in Section~\ref{sec:AdaFM}, we can insert the AdaFM layers after all convolution layers or just in the residual blocks (the same as BN and IN). Moreover, an alternative choice is to add AdaFM layers \textit{after} all activation layers. To evaluate the above three approaches, we conduct experiments for super resolution task $\times3\rightarrow\times4$ with filter size $5\times5$.
From the experimental results, we observe that adding AdaFM layers after activation is inferior to that before activation (32.00 dB, 31.84 dB evaluated on Set5 \cite{bevilacqua2012low}). The results of inserting AdaFM layers after all convolution layers and in the residual blocks make little difference (32.01 dB, 32.00 dB evaluated on Set5). To save computation, we insert AdaFM layers just in residual blocks before activation for all experiments.

\begin{table}[]
\small
\centering
\begin{tabular}{r|c|c}
\hline
\hline
Super resolution & SRResNet & basic model \\ 
\textbf{Set5}$\times$4& 32.05 & \textbf{32.13}\\
\hline
Denoising & DnCNN & basic model \\ 
\textbf{CBSD68} $\sigma$15 & 33.89 & \textbf{34.10}\\
\hline
DeJPEG & ARCNN & basic model \\
\textbf{LIVE1} $q10$ & 29.13 & \textbf{29.55}\\
\hline
\end{tabular}
\captionsetup{font={small}}
\vspace{-1em}
\caption{Comparisons with the state-of-the-art methods in PSNR.}
\label{tabel:basic}
\end{table}

\begin{figure}[]
	\vspace{-1.5em}
	\centering
	\includegraphics[scale=0.44]{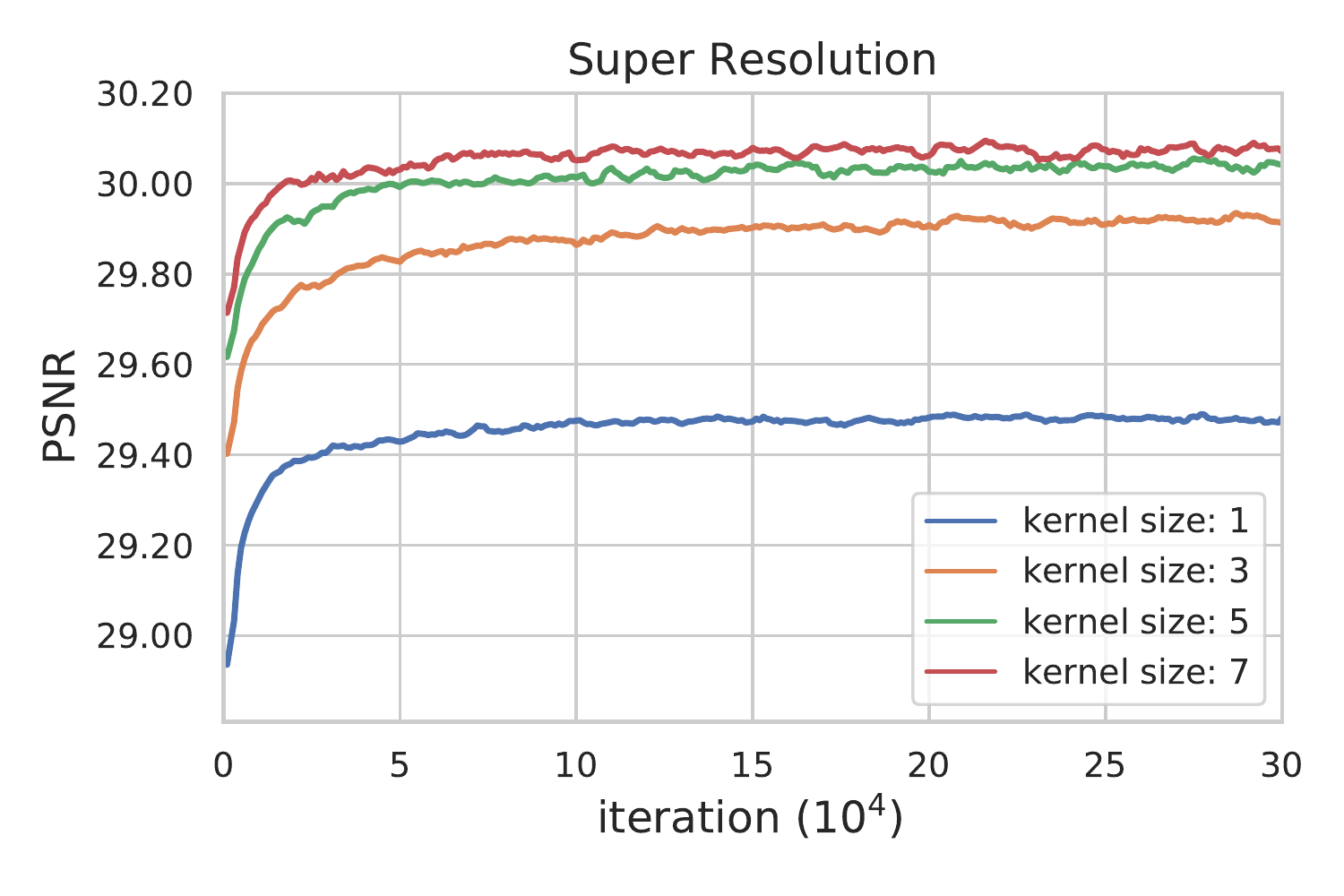}
	\vspace{-1.5em}
	\captionsetup{font={small}}
	\caption{The performances of adaptation with different filter sizes of AdaFM layers in super resolution on Set5 dataset.}
	\vspace{-2em}
	\label{kernel_size_figure}
\end{figure}

\textbf{Complexity analysis.}
We calculate the parameters of the basic model and AdaFM layers. Following previous works, we exclude the number of biases that perform add operation in network. The total parameters in basic model include the parameters of 16 residual blocks, 4 convolution layers and a pixelshuffle layer. As we insert the AdaFM layers in residual blocks, the number of AdaFM layers is equal to the number of convolution layers in residual blocks. Thus there are $16\times2\times64=2048$ filters in AdaFM layers. When the filter size is $1\times1$, $3\times3$, $5\times5$, the number of parameters is 2048, 18432, 51200, respectively, accounting for 0.15\% 1.31\% 3.65\% of the total parameters in the basic model. Note that these numbers are even smaller than the parameter number of a single residual block ($2\times64\times64\times9=73728$). Nevertheless, as AdaFM-Net is comparably larger than the basic model, we still need to verify whether it significantly improves the model capacity. In super resolution $\times4$, we train an AdaFM-Net with AdaFM layers of a large filter size $5\times5$ from scratch. The PSNR value on DIV2K (30.39 dB) is almost the same as that of the basic model (30.37 dB), indicating that the performance is not influenced by AdaFM layers. We can safely use the basic model as baseline to test the AdaFM-Nets. In another perspective, this also demonstrates the effectiveness of the proposed strategy, which adapts the model to different restoration levels with little additional computation cost.

\begin{table}[]
\small
\centering
\begin{tabular}{r|p{1.8em}p{1.8em}p{1.8em}p{2em}|p{2.5em}}
\hline
\hline
 PSNR(dB)& 1$\times$1 & 3$\times$3 & 5$\times$5 & 7$\times$7 & baseline  \\ \hline
SR \; \quad \quad \quad \quad \textbf{Set5} & 31.42 & 31.88 & 32.00 & \textbf{32.03} & 32.13 \\

\textbf{DIV2K100} & 29.89 & 30.20 & 30.28 & \textbf{30.30} & 30.37 \\
 \hline
DeJPEG \quad\;\textbf{LIVE1} & 29.35 & 29.39 & 29.41 & \textbf{29.42} & 29.55 \\
\hline
 
Denoising \textbf{CBSD68} & 26.35 & 26.38 & 26.39 & \textbf{26.40} & 26.49 \\
 \hline
\end{tabular}
\vspace{-1em}
\captionsetup{font={small}}
\caption{The PSNR results of adaptation with different kernel sizes of AdaFM layers in three tasks.}
\vspace{-1em}
\label{table:kernel_size}
\end{table}

\begin{table}[]
	\small
	\centering
	\begin{tabular}{r|ccc}
		\hline
		\hline
		PSNR(dB) & AdaBN & Conditional IN &  AdaFM-Net\\ 
		\hline
		\textbf{Set5}\,$\times3$ & 34.04 & 33.53 & \textbf{34.34}\\
		$\times4$ & 28.70 & 31.30 & \textbf{32.00} \\
		\hline
		\textbf{LIVE1}\,$q80$ & 38.29 & 36.99 & \textbf{38.81}\\
		$q10$ & 27.61 & 28.89 & \textbf{29.35}\\
		\hline
		\textbf{CBSD68} $\sigma15$ & 33.83 & 31.33 & \textbf{34.10}\\
		$\sigma75$ & 19.68 & 24.15 & \textbf{26.35} \\
		\hline
	\end{tabular}
	\vspace{-1em}
	\captionsetup{font={small}}
	\caption{Comparisons with AdaBN \cite{Li2017RevisitingBN} and conditional IN \cite{dumoulin2017learned}}
	\vspace{-2em}
	\label{comparison}
\end{table}

\begin{table*}[t]\centering
\renewcommand{\arraystretch}{0.9}
\vspace{-1.5em}
\setlength{\tabcolsep}{3pt}
\begin{tabular}{@{}rccccccccccc@{}}
\multicolumn{12}{c}{\textbf{Adaptation in Super Resolution}}\\\specialrule{0em}{0pt}{2pt}
\toprule
& \multicolumn{4}{c}{$\textit{\textbf{range1}}$} & & \multicolumn{3}{c}{$\textit{\textbf{range2}}$} &
& \multicolumn{2}{c}{$\textit{\textbf{direction}}$}\\
\cmidrule{2-5} \cmidrule{7-9} \cmidrule{11-12}
& $\times2\rightarrow\times3$ & $\times3\rightarrow\times4 $& $\times4\rightarrow\times5 $& $\times5\rightarrow\times6$ & & $\times2\Rightarrow\times4$ & $\times3\Rightarrow\times5$ & $\times4\Rightarrow\times6$ & & $\times3\leftarrow\times4$ & $\times2\Leftarrow\times4$\\ \midrule
\textbf{Set5} & 34.34 & 32.13 & 30.26 & 28.74 && 32.13 & 30.26 & 28.74 && 34.34 & 37.84\\
AdaFM-Net & 33.98 & 32.00 & 30.16 & 28.73 && 31.66 & 29.98 & 28.61 && 34.11 & 37.11\\
\cmidrule{2-12}
PSNR distance  & 0.36 & \textbf{0.13} & \textbf{0.10} & \textbf{0.01} && 0.47 & 0.28 & \textbf{0.13} && 0.23 & 0.73\\\midrule

\textbf{DIV2K100} & 32.35 & 30.37 & 29.04 & 28.10 && 30.37 & 29.04 & 28.10 && 32.35 & 36.00\\
 AdaFM-Net & 32.07 & 30.28 & 29.02 & 28.09 && 30.01 & 28,88 & 28.00 && 32.14 & 35.13\\
\cmidrule{2-12}
  PSNR distance    & 0.28 & \textbf{0.09} & \textbf{0.02} & \textbf{0.01} && 0.36 & \textbf{0.16} & \textbf{0.10} && 0.21 & 0.87\\ 
\bottomrule
\end{tabular}
\vspace{-1em}
\caption{Adaptation results. The PSNR distances within 0.2 dB are shown in bold.}
\vspace{-1.5em}
\label{sr_range_direction}
\end{table*}

\subsection{Evaluation of Model Training}
In this section, we evaluate our proposed method on three image restoration tasks, super resolution, denoising, and compression artifacts reduction (JPEG Deblocking or DeJPEG). The basic settings are shown below. 

For super-resolution, we train our models in RGB channels and calculate the PSNR in y-channel on two widely used benchmark datasets -- Set5~\cite{bevilacqua2012low} and the test set of DIV2K~\cite{agustsson2017ntire}. We evaluate our methods on upscaling factors $\times2, \times3, \times4, \times5, \times6$. All other settings remain the same as SRCNN~\cite{dong2014learning}. In denoising, we use Gaussian noise and consider 5 noise levels, i.e., $\sigma=15, 25, 35, 50, 75$. Following DnCNN~\cite{zhang2017beyond}, the models are trained with RGB channels and evaluated in RGB channels on CSBD68~\cite{roth2005fields} dataset. For DeJPEG, we use the JPEG quality $q = 80, 60, 40, 20, 10$ in MATLAB JPEG encoder. Similar as ARCNN~\cite{Dong_2015_ICCV}, our models are trained and tested in y channel only. LIVE1 \cite{1709988} dataset is used for evaluation.

\textbf{Filter Size.}
First, we need to determine the filter size of AdaFM layers for different problems. We denote the adaptation from the start level $L_a$ to the end level $L_b$ as $L_a\rightarrow L_b$. The basic model is trained on $L_a$, and AdaFM-Net is tested on $L_b$.

For the super resolution task $\times3\rightarrow\times4$, we compare the performance of AdaFM-Net with various filter sizes -- 1$\times$1, 3$\times$3, 5$\times$5 and 7$\times$7. The convergence curves on Set5 are plotted in Figure~\ref{kernel_size_figure} , and the quantitative results are presented in Table \ref{table:kernel_size}. In general, larger filters can achieve better performance. Notably, the PSNR gap between $1\times 1$ and $3\times 3$ is larger than 0.4 dB. However, this trend does not always hold when the filter size is expanded to 7$\times$7. Therefore, we use the filter size 5$\times$5 to conduct the following experiments for the super resolution tasks. 

Similar as in super resolution, we compare the performance with different filter sizes (1$\times$1, 3$\times$3, 5$\times$5 and 7$\times$7) for denoising task $\sigma15\rightarrow \sigma75$ and DeJPEG task $q80\rightarrow q10$. Results shown in Table \ref{table:kernel_size} indicate that in both two tasks, filter size 1$\times$1 can already achieve excellent performance. The PSNR gap between $1\times 1$ and $7\times 7$ is less than 0.1 dB. Considering the computation cost, we use filter size $1\times1$ for all denoising and DeJPEG experiments.

\begin{table}[]
	\small
	\setlength{\tabcolsep}{5pt}
	\renewcommand{\arraystretch}{0.9}
	\begin{tabular}{@{}rccccc@{}}
		\toprule
		& \multicolumn{4}{c}{$\textit{\textbf{range}}$} 
		& \multicolumn{1}{c}{$\textit{\textbf{direction}}$}\\
		\cmidrule(r{4pt}){2-5}  \cmidrule(l{4pt}){6-6}
		\textbf{\textit{DeJPEG}}
		& 80$\rightarrow$60 & 80$\rightarrow$40 & 80$\rightarrow$20 & 80$\rightarrow$10 &80$\leftarrow$10 \\ \midrule
		\textbf{LIVE1} & 36.00  &34.34 &31.93 &  29.55 &  38.81\\
		AdaFM-Net & 35.98 & 34.29 &31.81 & 29.35 &37.77\\
		\cmidrule{2-6}
		distance & \textbf{0.02} &\textbf{ 0.05} &\textbf{ 0.12} & \textbf{0.20} &1.04 \\
		\bottomrule
		\\
		\specialrule{0em}{0pt}{-0.5em}
		\textit{\textbf{Denoising}}
		& 15$\rightarrow$25 & 15$\rightarrow$35 & 15$\rightarrow$50 & 15$\rightarrow$75 & 15$\leftarrow$75 \\ \midrule
		\textbf{CBSD68} & 31.44 & 29.82 & 28.20 & 26.49 & 34.10\\
		AdaFM-Net & 31.43 & 29.78 & 28.13 & 26.35 & 33.42\\
		\cmidrule{2-6}
		distance & \textbf{0.01} & \textbf{0.04} & \textbf{0.07} & \textbf{0.14} & 0.68 \\
		\bottomrule
	\end{tabular}
	\vspace{-1em}
	\captionsetup{font={small}}
	\caption{Adaptation results of DeJPEG and Denoising. The PSNR distances within 0.2 dB are shown in bold.}
	\vspace{-2em}
	\label{jpeg_range_direction}
\end{table}

\textbf{Direction.}
\label{sec:direction}
The second step is to find the best adaptation direction. Before experiments, it is essential to clarify the way of measurement. For task $ L_a\rightarrow L_b$, the baseline is the basic model trained on $L_b$ with performance $P_{bas}$, and the AdaFM-Net is finetuned on $L_b$ with performance $P_{ada}$. Then the PSNR distance $|P_{bas}- P_{ada}|$ is used to evaluate the \textit{adaptation accuracy} of AdaFM-Net. In experience, 0.3 dB is regarded as a significant PSNR gap in image restoration. In other words, if the distance $|P_{bas}- P_{ada}|$ exceeds 0.3 dB, then the adaptation is NOT well-suited for applications. 

We conduct three pairs of experiments -- super resolution task $\times3\rightarrow\times4$ and $ \times4\rightarrow \times3$, denoising task $\sigma15\rightarrow \sigma75$ and $\sigma75\rightarrow \sigma15$, DeJPEG task $q80\rightarrow q10$ and $q10\rightarrow q80$. Results are shown in Table~\ref{sr_range_direction}, \ref{jpeg_range_direction}. In all three problems, the tasks with direction from easy to hard (i.e., $\times3\rightarrow \times4$, $\sigma15\rightarrow \sigma75$, $q80\rightarrow q10$) achieve better adaptation results. Here, \textit{easy} and \textit{hard} refer to the difficulty of restoring the input images. For example, in DeJPEG, the PSNR distance of $q80\rightarrow q10$ is 0.2 dB, which is much lower than that of the inverse direction $q10\rightarrow q80$ -- 1.04 dB.

\textbf{Range.}
In this subsection, we investigate the influence of the adaptation range. Generally, by fixing the start level $L_a$, we change the end level $L_b$ and test the adaptation accuracy. 

Different from previous sections, we start discussion with denoising and DeJPEG, where the trend of range is more obvious. In denoising, we start with $\sigma15$ and change the end level from $\sigma25$ to $\sigma75$. In DeJPEG, we start with $q80$ and change the end level from $q60$ to $q10$. The adaptation results are shown in Table~\ref{jpeg_range_direction}. It is observed that better adaptation accuracy is obtained with a smaller range.
In addition, the proposed AdaFM can easily handle very large range in either denoising or DeJPEG.

For super resolution, we find it hard to adapt the model across even 2 upscaling factors. For example, in Table~\ref{sr_range_direction}, the PSNR distance for the task $\times2\rightarrow \times4$ exceeds 0.3 dB for all three test sets, indicating that we should not further enlarge the range to 3. When fixing the range to be 1 and 2 upscaling factors, we change both the start and end level to see the change of results. From Table~\ref{sr_range_direction}, we can conclude that the adaptation is easier (lower PSNR distance) with a harder start level (e.g., $\times4\rightarrow \times5$ is better than $\times3\rightarrow \times4$).

\begin{figure*}
	\centering
	\vspace{-2em}
	\includegraphics[scale=0.27]{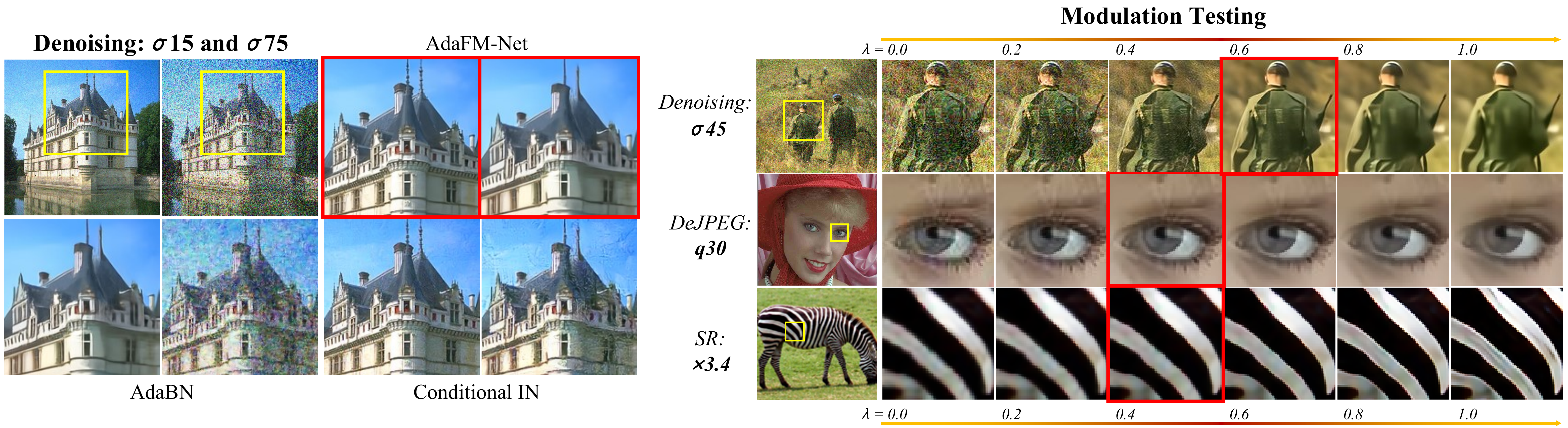}
	\captionsetup{font={small}}
	\vspace{-1em}
	\caption{Left: Artifacts on the output images produced by AdaBN and conditional instance normalization. Right: Modulation testing in Denoising (CBSD68), DeJEPG (LIVE1) and Super Resolution (Set14 \cite{zeyde2010single}).}
	\vspace{-1.5em}
	\label{fig:baselines}
\end{figure*}

\vspace{-0.7em}
\subsubsection{Comparison with AdaBN and Conditional IN}
\quad We compare with state-of-the-art methods on super resolution task $\times3\rightarrow \times4$, denoising task $\sigma15\rightarrow \sigma75$ and DeJPEG task $q80\rightarrow q10$. To compare with AdaBN \cite{Li2017RevisitingBN}, we train a network with batch normalization after all convolutional layers in the residual blocks, and then change all the statistics in BN layers during testing. We also use conditional IN \cite{huang2017arbitrary} to handle different levels of restoration. The results are shown in Table \ref{comparison}.
It can be obviously observed that neither of the two methods can obtain reasonable super resolution, denoising and DeJPEG results. Therefore, they are not suitable for image restoration tasks. Qualitative comparisons are also shown in Figure~\ref{fig:baselines}, where we observe clear artifacts on their output images.

\subsection{Evaluation of Modulation Testing}
\label{sec:modulation}
In modulation testing, continuously manipulating the interpolation coefficient $\lambda$ could gradually change the output effect. If the input image is fixed, then the output image will become sharper or smoother with the increase of $\lambda$, as shown in Figure~\ref{fig:baselines}. On the other hand, we can choose different $\lambda$ to deal with different kinds of degraded images. As presented in Section~\ref{sec:test}, the coefficient $\lambda$ can be formulated as a polynomial function of restoration level $L_c$ -- $\lambda =\sum_{j=0}^M w_j L_c ^j$. In this subsection, we investigate the curving fitting with different ranges in DeJPEG problem. Similar investigations on super resolution and denoising problems can be found in supplementary file. 

\begin{figure}[h]
	\vspace{-1em}
	\centering
	\includegraphics[scale=0.32]{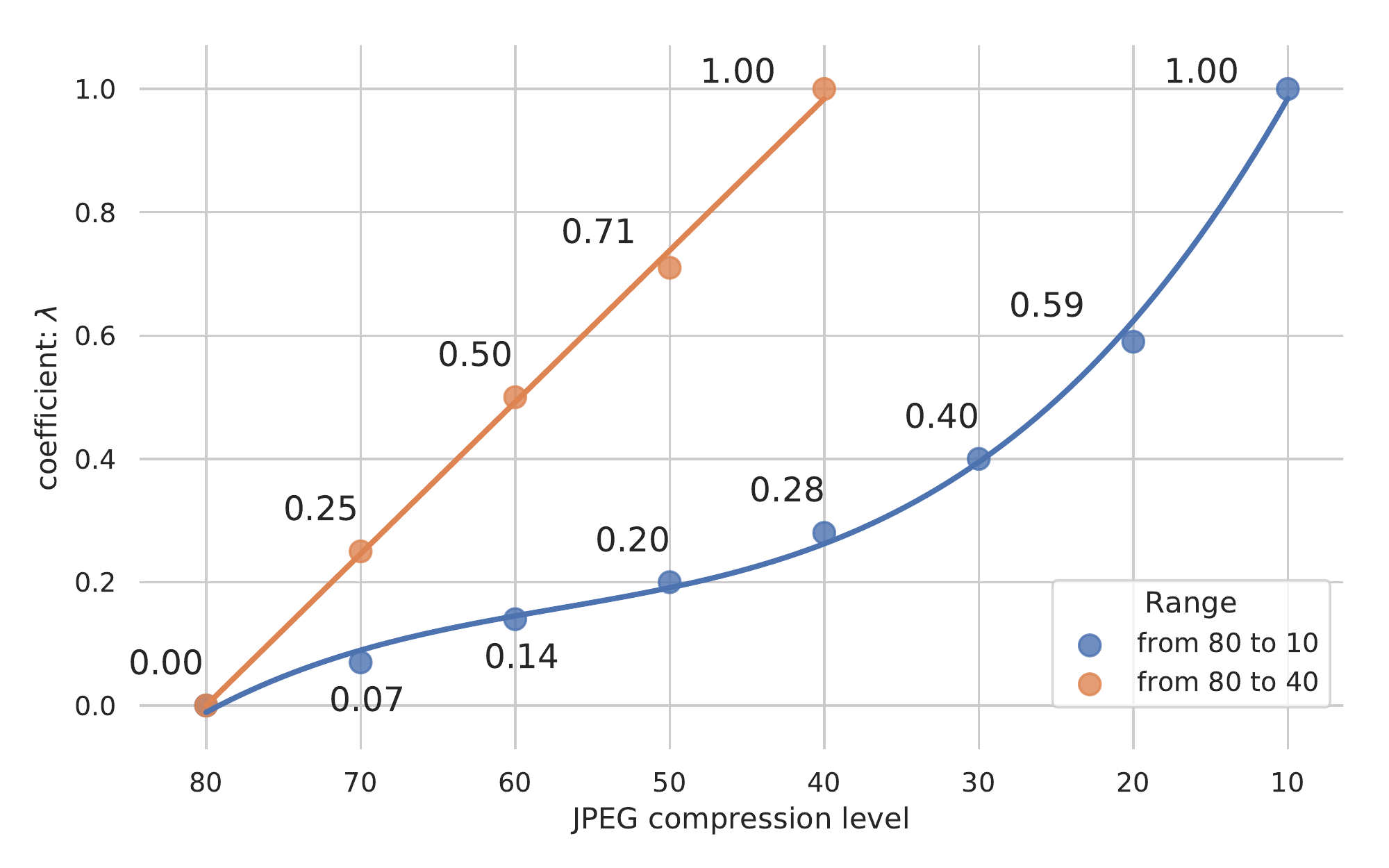}
	\includegraphics[scale=0.32]{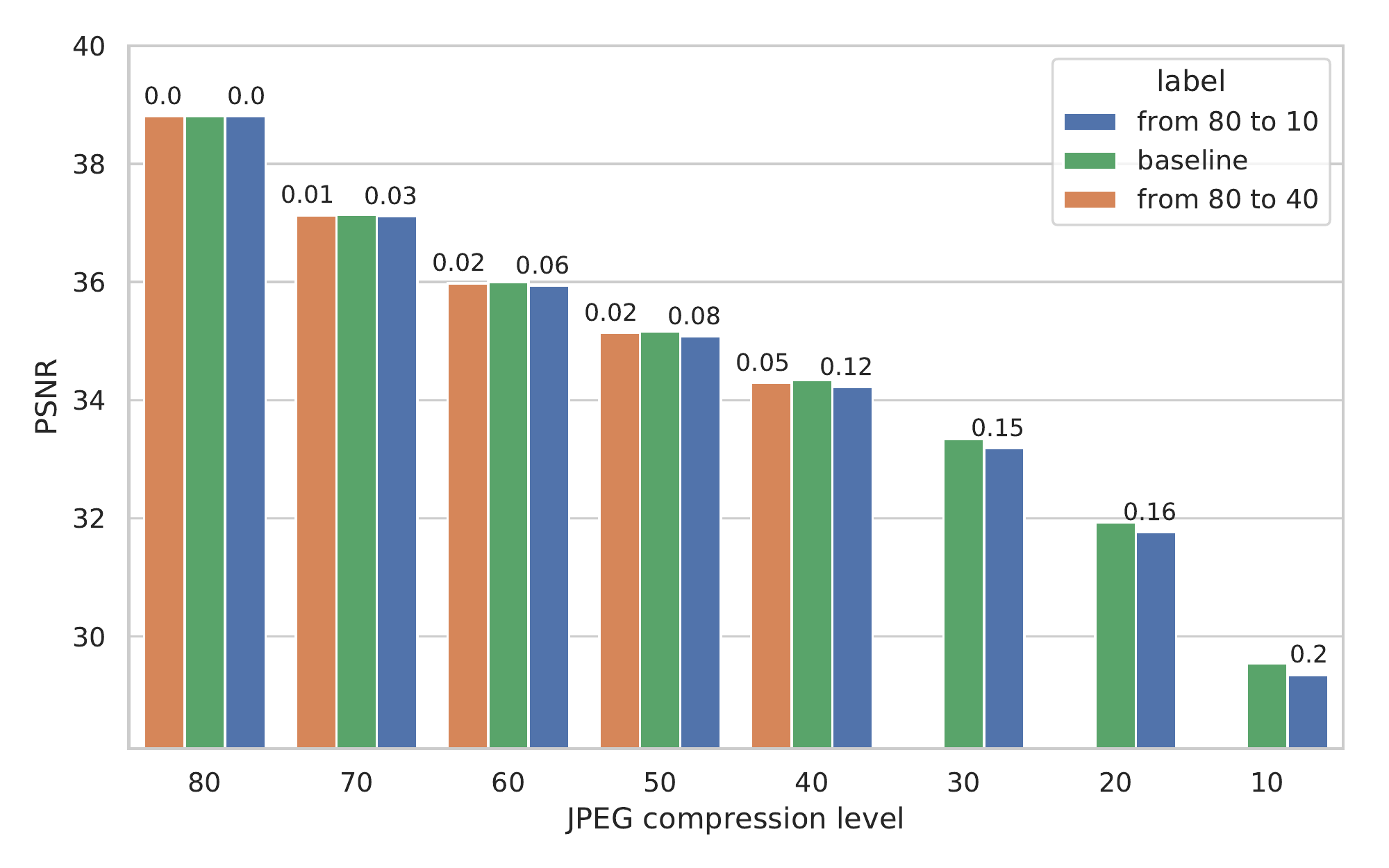}
	\vspace{-1em}
	\captionsetup{font={small}}
	\caption{Top: the curve fitting with different ranges in DeJPEG problem; Bottom: the value of PSNR distance is annotated above each bar.}
	\label{fig:modulation}
	\vspace{-2em}
\end{figure}

We first investigate the DeJPEG task $ q80\rightarrow q10$.
We select 6 middle levels -- $L_c=q70,60,50,40,30,20$ -- between $ q80$ and $ q10$. Then for a given level $L_c$, we use the test images of $L_c$ as inputs, and adjust $\lambda$ to obtain different outputs of AdaFM-Net. After calculating the PSNR values on LIVE1 test set, we select the $\lambda$ that achieves the best PSNR as the best coefficient. For example, see the blue line in Figure~\ref{fig:modulation}, the best coefficient for level  $q60$ and $q30$ are 0.14, 0.40, respectively. After we have obtained all middle points, we fit the curve by a cubic function:$\lambda=1.51-6.24\times10^{-2}L_c+1.01\times10^{-3}L_c^2-5.91\times10^{-6}L_c^3$. Then for arbitrary levels between $q80$ and $q10$, we can use this function to predict its corresponding interpolation coefficient. If we test a smaller range, such as $q80\rightarrow q50$, then a simple straight line could fairly connect all middle points (see the orange line in Figure~\ref{fig:modulation}). In other words, the polynomial function is linear. This property holds for smaller ranges such as $q80\rightarrow q60$.

To verify whether the interpolated image is of high quality, we use the PSNR distance on LIVE1 test set as the evaluation metric. Specifically, the basic model trained on level $L_c$ is used as the baseline, and the PSNR distance is calculated between the PSNR of AdaFM-Net and that of a well-trained baseline model. The smaller of the PSNR distance the better of the adaptation/modulation accuracy. Figure~\ref{fig:modulation} illustrates the PSNR distances in two DeJPEG tasks. It is observed that all PSNR distances are below 0.2 dB, indicating that the output quality is good enough for practical usages. Further, the PNSR distances of the small-range task $q80\rightarrow q50$ is much lower than the large-range task $q80\rightarrow q10$. Thus modulation across smaller ranges achieves better performance. For higher request of modulation quality, we can decompose a large range to several small ranges, and train AdaFM-Nets for each sub-task. We can balance the performance and computation burden according to different applications.

\section{Conclusion}
We present a method that allows continual modulation of restoration levels in a single CNN for versatile and flexible image restoration. 
The core idea of our method is to handle images with arbitrary degradation levels with a single model, which consists of a basic model and a modulation layer -- AdaFM layer. We further propose the learning and modulating strategies of the AdaFM layers. In test time, the model can be adapted to any restoration level by directly adjusting the AdaFM layers without an additional training stage.

\textbf{Acknowledgements}. 
This work is partially supported by National Key Research and Development Program of China (2016YFC1400704), Shenzhen Research Program (JCYJ20170818164704758, JCYJ20150925163005055, CXB201104220032A), and Joint Lab of CAS-HK.

{\small
\bibliographystyle{ieee_fullname}

}

\end{document}